\documentclass[conference]{IEEEtran}
\IEEEoverridecommandlockouts

\usepackage{tikz}
\usepackage{textcomp}
\usepackage{hyperref}
\usepackage{lipsum}

\usepackage{gensymb}
\usepackage[utf8]{inputenc}
\usepackage[T1]{fontenc}
\ifCLASSINFOpdf
   \usepackage{graphicx}
  \DeclareGraphicsExtensions{.jpeg,.png}
\else
\fi
\usepackage{amsmath}
\usepackage[symbol]{footmisc}
\usepackage{array}
\usepackage{subcaption}
\usepackage{authblk}
\begin{document}
\title{Auto-Detection of Tibial Plateau Angle in Canine Radiographs Using a Deep Learning Approach}

\author[1,2]{Masuda Akter Tonima}
\author[1]{F M Anim Hossain}
\author[2]{Austin DeHart}
\author[1,*]{Youmin Zhang}
\affil[1]{Gina Cody School of Engineering, Concordia University, Montreal, Quebec, Canada}
\affil[2]{Innotech Medical Industries Corp., Vancouver, British Columbia, Canada}
\affil[*]{youmin.zhang@concordia.ca}

\renewcommand\Authands{ and }
\maketitle
\begin{abstract}
Stifle joint issues are a major cause of lameness in dogs and it can be a significant marker for various forms of diseases or injuries. A known Tibial Plateau Angle (TPA) helps in the reduction of the diagnosis time of the cause. With the state of the art object detection algorithm YOLO, and its variants, this paper delves into identifying joints, their centroids and other regions of interest to draw multiple line axes and finally calculating the TPA. The methods investigated predicts successfully the TPA within the normal range for 80 percent of the images.
\end{abstract}
\IEEEpeerreviewmaketitle
\section{Introduction}
\IEEEPARstart \noindent Canine stifle joint issues are one of the oldest and most faced issues in the veterinary orthopedics sector \cite{seo2020measurement}. The value of Tibial Plateau Angle (TPA) can be used to identify issues that exist in a canine's leg, how it may react to different forms of injuries or even if there is presence of pre-existing conditions \cite{sorensson2021evaluation}. Injuries such as the cranial cruciate ligament (CCL) rupture (CCLR) are very common, major and progressively deteriorate stifle joints of canines permanently; automated Tibial Plateau Angle (TPA) assessment can help in shortening the process of CCL detection during surgical procedures \cite{pacchiana2003surgical}, additionally, as the TPA value plays an important role in the distribution of force when a dog walks, it can also determine whether or not there is excessive cranial tibial thrust that may predispose canines to CCLR \cite{seo2020measurement, kim2015measurement}.

\indent The utility of automation has become ubiquitous in the modern world where everything is electronically powered and we, humans rely more and more on artificial intelligence for assistance. Automation in image annotation has become a major tool in the medical field, driving patient care decisions \cite{grady2020system}. The latest method of obtaining such decisions in an automated manner involves using a processor for digital image representation acquisition that simultaneously generates annotations and determines the associations between multiple annotations of objects of same class or group \cite{grady2020system}. This same processor simultaneously works on determining and representing the said classes while recording the annotations in its limited memory \cite{grady2020system}. There has been variety of work regarding determination TPA considering multiple factors such as age, sex, breed etc. as seen in studies \cite{kim2015measurement, pacchiana2003surgical, sorensson2021evaluation, moore1995cranial}, however none thus far for the automation of the process. This study delves into this aspect of application and completes the first stage of automation in TPA determination. 
\section{TPA Calculation}
\begin{figure}[b]
    \centering
    \begin{subfigure}[b]{0.26\linewidth}
      \centering
      \includegraphics[width=0.8in, height=1.5in]{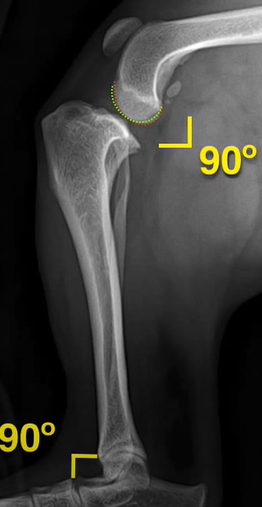}
      \caption{}
    \end{subfigure}
    \begin{subfigure}[b]{0.26\linewidth}
      \centering
      \includegraphics[width=0.8in, height=1.5in]{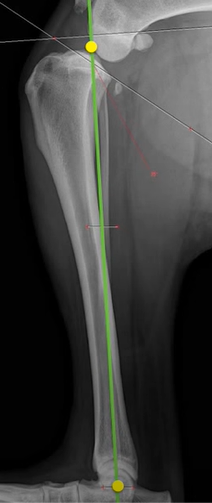}
      \caption{}
    \end{subfigure}
    \begin{subfigure}[b]{0.26\linewidth}
      \centering
      \includegraphics[width=0.8 in, height=1.5in]{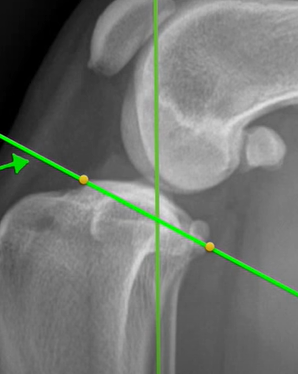}
      \caption{}
    \end{subfigure}
   \caption{\small{TPA requirements (a) Stifle (knee joint) and Tarsus (ankle joint) flex, (b) Green line representing the FTL and (c) MTPL pointed with green arrow \cite{MTPAvid}.}}
   \label{fig:tparecs}
\end{figure}

\noindent TPA calculation has some prerequisites that must be fulfilled, these are:
\begin{enumerate}
   \item The Stifle and Tarsus must be 90\degree flexed;
   \item Functional Tibial line must be formed by connecting
   \begin{enumerate}
     \item Centre of Talus
     \item The centre of intercondylar eminences shown in Fig. \ref{fig:tparecs};
   \end{enumerate}
   \item The Medial Tibial Plateau line should be drawn using the first two.
 \end{enumerate}

\begin{figure}[t]
    \centering
    \begin{subfigure}[t]{0.3\linewidth}
      \centering
      \includegraphics[width=0.9in,height=1in]{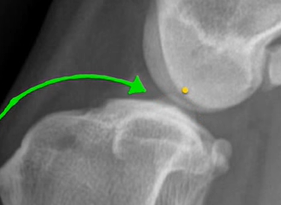}
      \caption{}
    \end{subfigure}
    \begin{subfigure}[t]{0.3\linewidth}
      \centering
      \includegraphics[width=0.9in,height=1in]{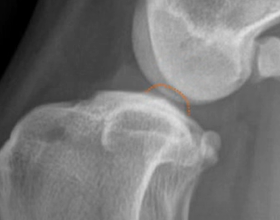}
      \caption{}
    \end{subfigure}
    \begin{subfigure}[t]{0.3\linewidth}
      \centering
      \includegraphics[width=0.9in,height=1in]{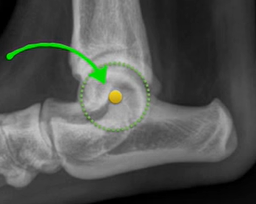}
      \caption{}
    \end{subfigure}
   \caption{\small{Functional Tibial line drawing needs; (a) Intercondylar Eminence point, (b) Intercondylar Eminence point reference and (c) Centre of Talus \cite{MTPAvid}.}}
   \label{fig:ftlrecs}
\end{figure}
\begin{figure}[b]
\centering
\includegraphics[height=2in]{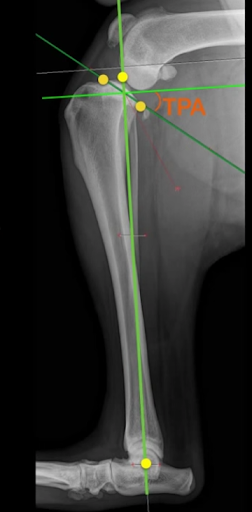}
\caption{\small{Tibial Plateau Angle shown in relation with the new perpendicular line drawn and the MTPL \cite{MTPAvid}.}}
\label{fig:tparef}
\end{figure}

\indent Following the identification of points of interests, and drawing of the lines of interest, i.e. the Functional Tibial Line (FTL) and the Medial Tibial Plateau Line (MTPL), another line is drawn such that the relation of new line and the FTL is 90\degree. Tibial plateau angle is the angle between this new line and the MTPL; this is shown in Fig. \ref{fig:tparef}.
Normal TPA values can range from 18 to 25 degrees \cite{seo2020measurement}; the large range can be attributed to the large range of breed, body weight, age etc.

\subsection{Training Dataset}
\begin{figure}[t]
    \centering
    \includegraphics[width=0.3\columnwidth]{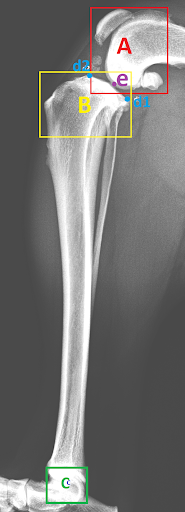}
    \caption{\small{Example of manual annotations, highlighting the regions of interests \cite{labelIMg}.}}\label{fig:tpaann}
\end{figure}

\begin{figure}[t]
    \centering
    \includegraphics[width=.3\columnwidth, height=1.5in]{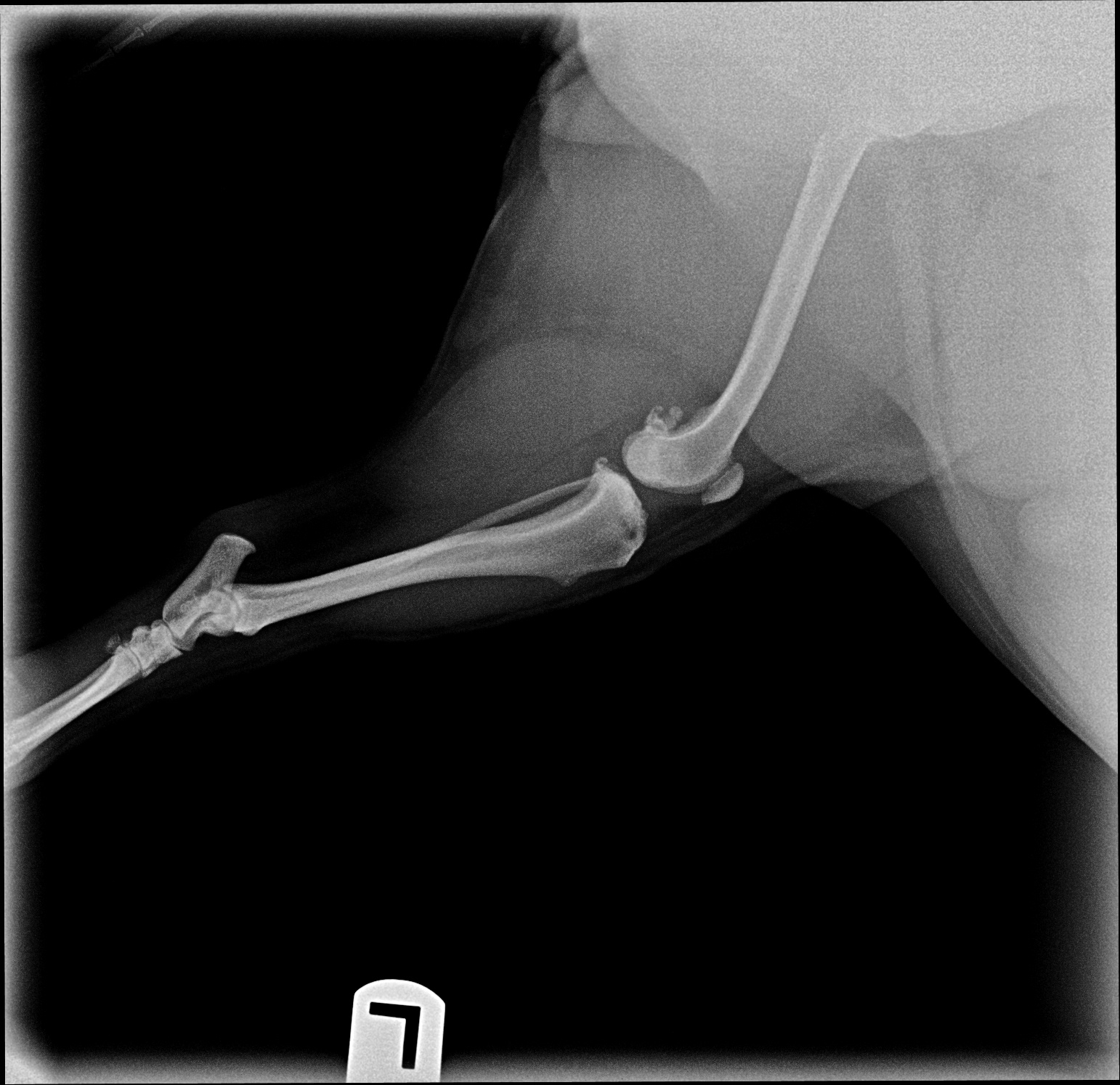}\hfill
    \includegraphics[width=.3\columnwidth, height=1.5in]{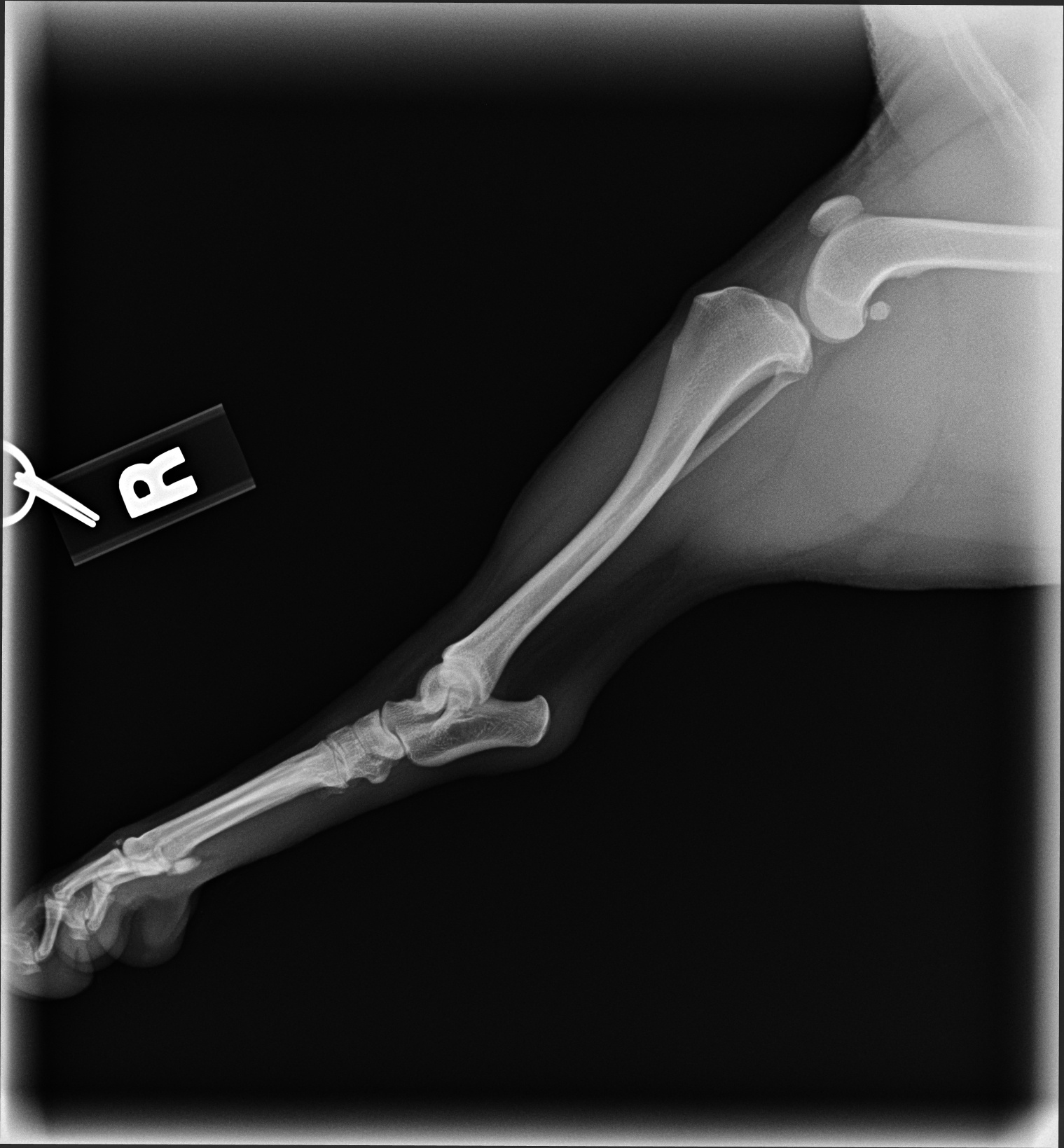}\hfill
    \includegraphics[width=.3\columnwidth, height=1.5in]{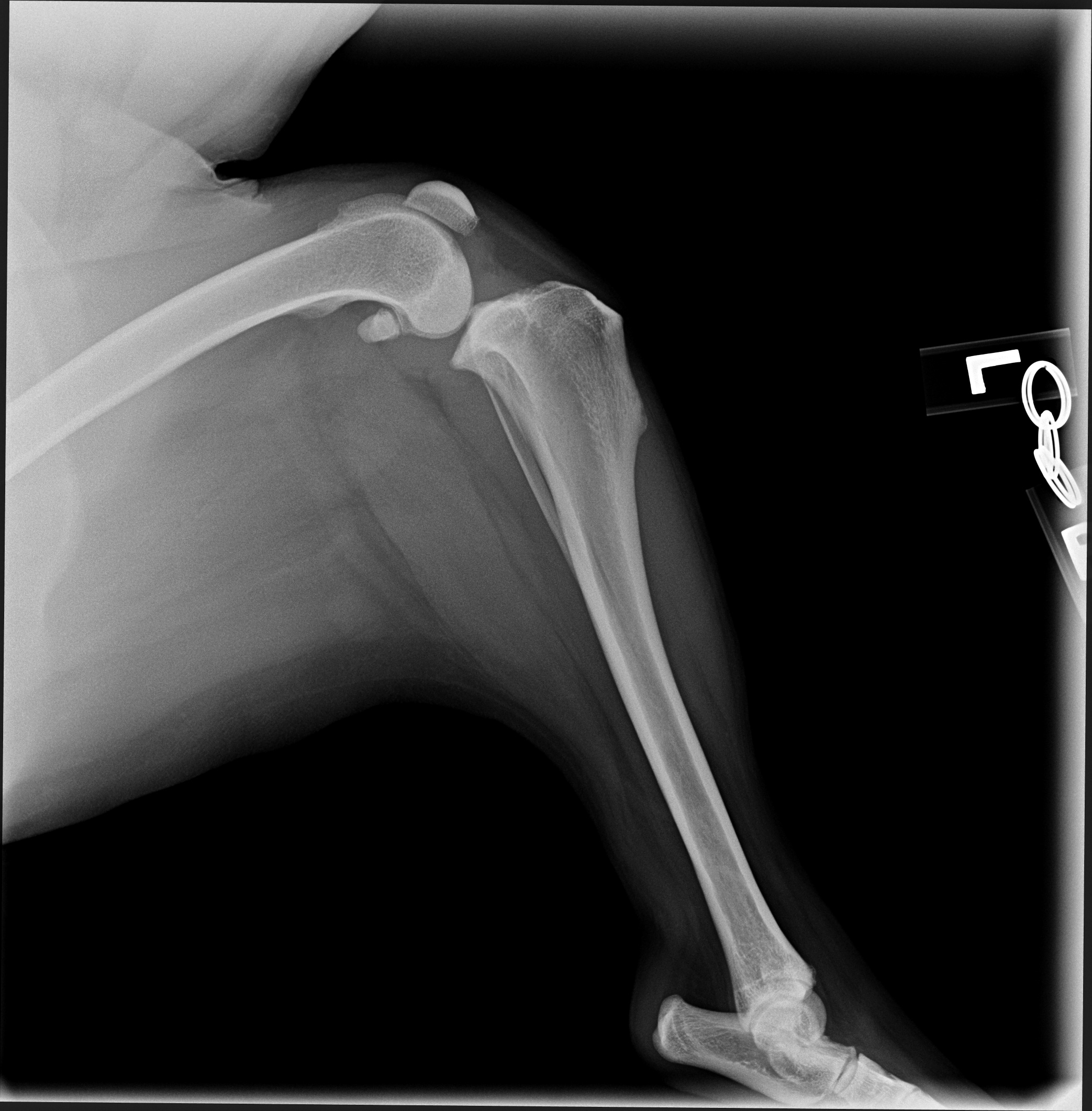}\hfill
    \\[\smallskipamount]
    \includegraphics[width=.3\columnwidth, height=1.5in]{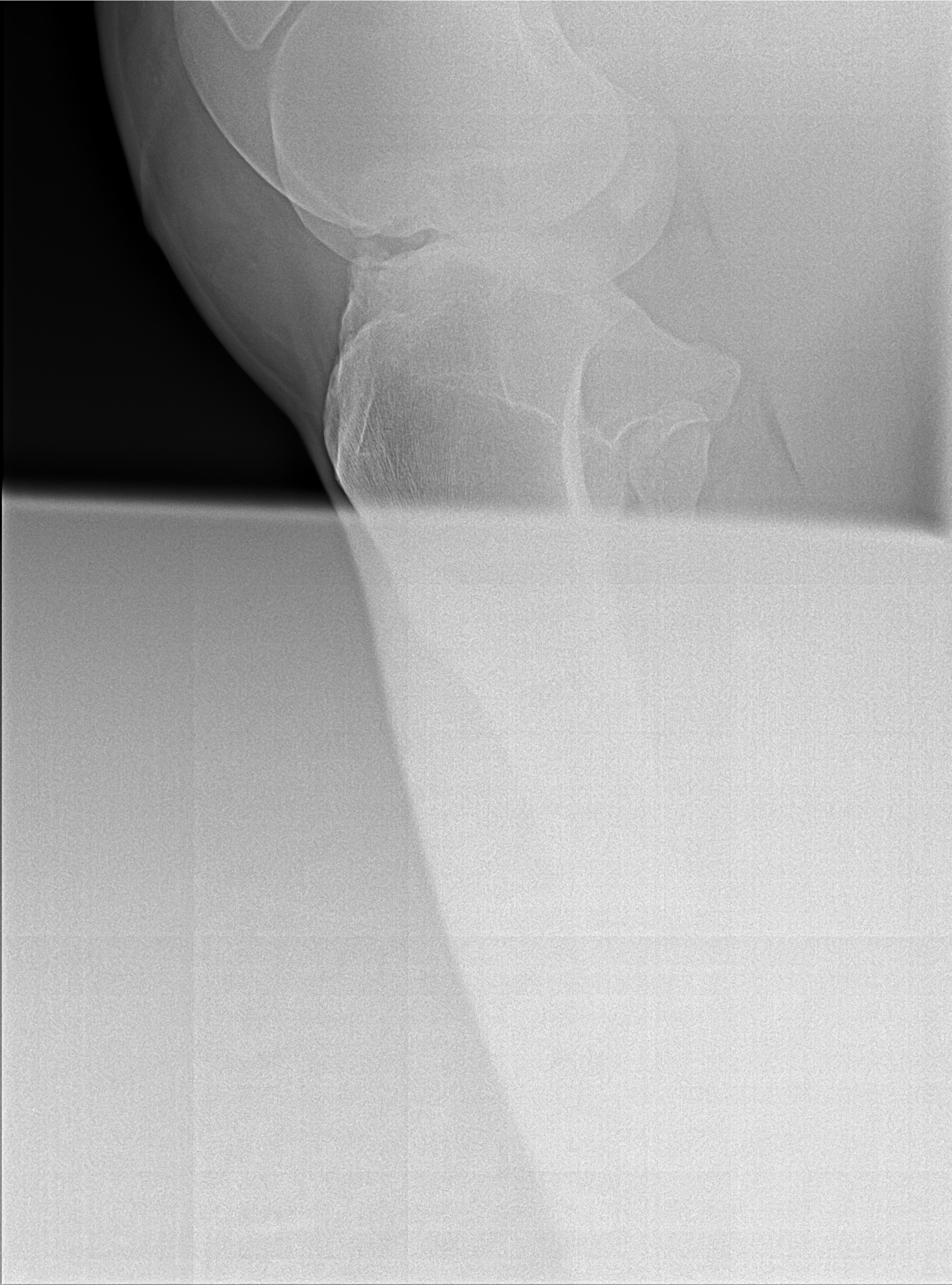}\hfill
    \includegraphics[width=.3\columnwidth, height=1.5in]{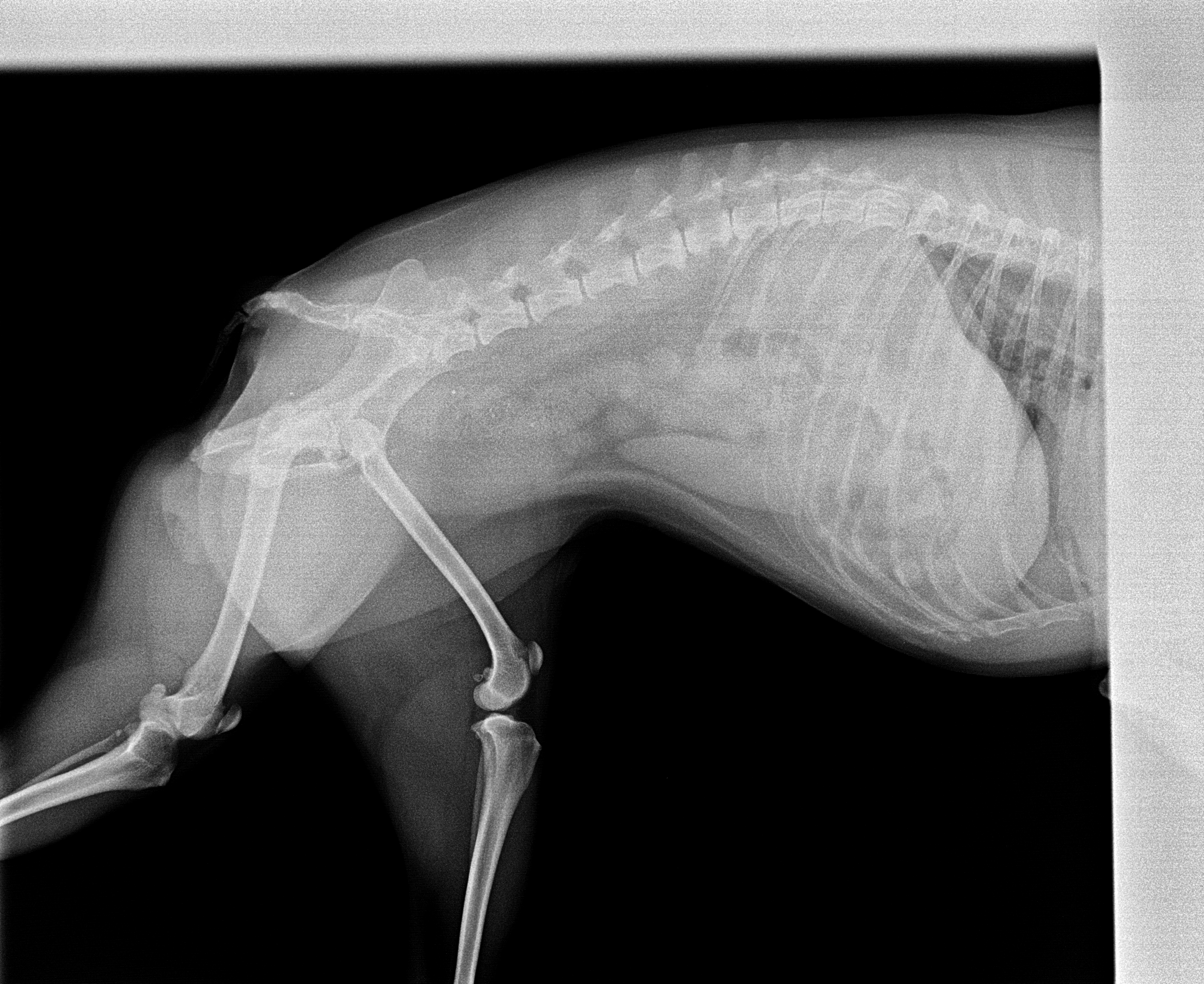}\hfill
    \includegraphics[width=.3\columnwidth, height=1.5in]{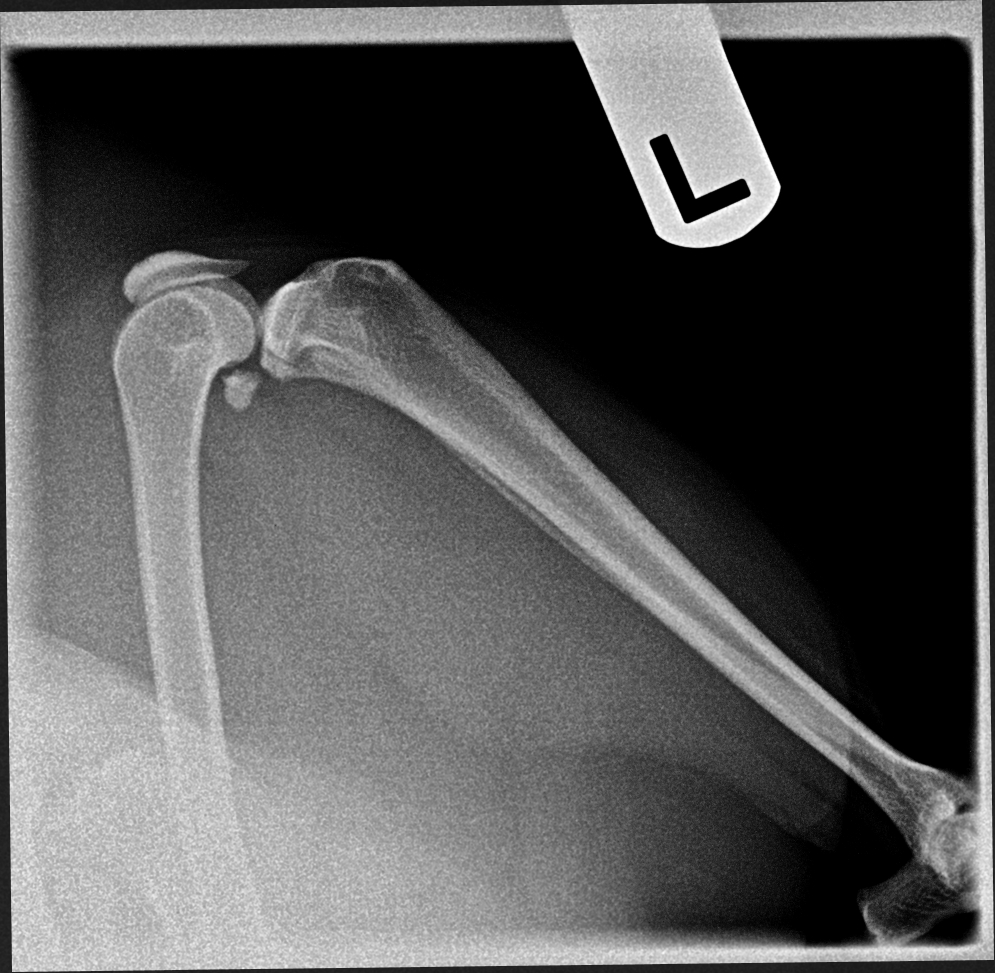}\hfill
    \caption{\small{Some examples of difficult data}}\label{fig:difdat}
\end{figure}

\noindent The dataset used for training was collected from various veterinary clinics, as none is available in the public domain, and the objects of interests were manually annotated. The resolution of these images vary largely as they are collected from various sources, thus for the sake of uniformity the images are all scaled to fit the same dimensions. The first part of this project was to develop a lightweight radiograph image sorting algorithm reported in \cite{Tonima_2021}. The images chosen for the task described in this manuscript were all classified to be lateral lower body images by that sorting algorithm. Examples of manual annotations of objects of interests are given in Fig. \ref{fig:tpaann}: here regions A, B and C identify the joints while the point 'e' identifies the centre of Talus and regions d1 and d2 identify the points that form the MTPL.
\\\indent The images from the source dataset had issues and needed to be consolidated into a more usable framework. These issues were mostly due to inconsistencies in practices of radiographers and movement by the animals during radiography that resulted in radiographs with incomplete data, inconsistent image quality, such as varying contrast, brightness and positioning of point of interest or images that failed to meet the prerequisite conditions for this task. Examples of difficult data are shown in Fig. \ref{fig:difdat}.
\\\indent Most of the source dataset of over a thousand unique knee radiographs were distorted, had poor patient positioning, or were otherwise unfit for annotating. 250 of the original images were of sufficient quality to contribute to this effort. These images were set to have 6 different classes for training purposes, as mentioned and shown in Fig. \ref{fig:tpaann}. These were then trained using YOLOv3 \cite{redmon2018YOLOv3} and the results of the predicted annotations are shown in Table \ref{table:tpares}. From these predictions the centroids were extracted, which is then used to plot the FTL and MTPL. Then using the method mentioned in Section II, TPA is calculated.
Examples of images, the region of interests detection and their respective TPA determination is shown in Fig. \ref{fig:tpapred} and Table \ref{table:tpares}:
\begin{figure}[ht!]
    \centering
    \begin{subfigure}[]{0.4\linewidth}
    \centering
    \includegraphics[width=1.2in, height=2in]{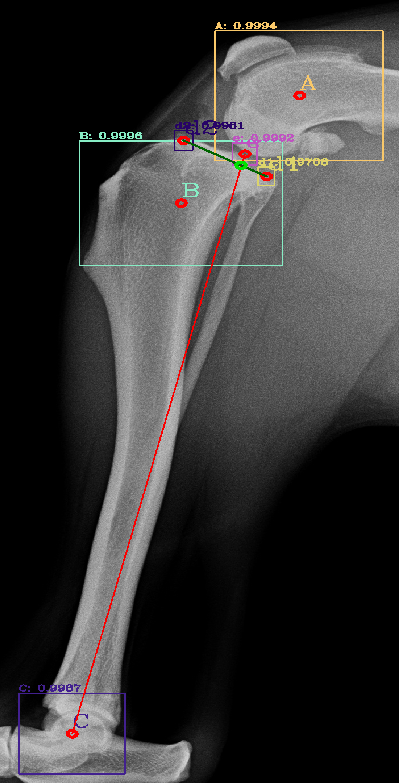}\hfill
    \caption{}
    \end{subfigure}
    \begin{subfigure}[]{0.4\linewidth}
      \centering
      \includegraphics[width=1.2in, height=2in]{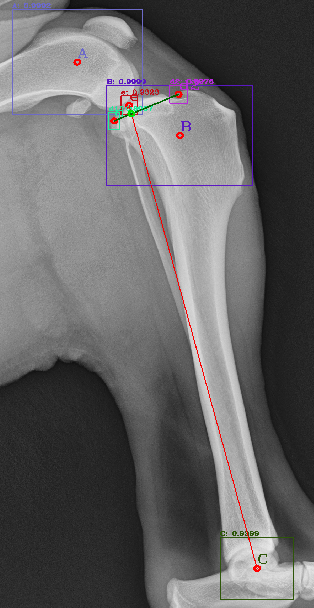}\hfill
    \caption{}
    \end{subfigure}
    \\[\smallskipamount]
    \begin{subfigure}[]{0.4\linewidth}
      \centering
      \includegraphics[width=1.2in, height=2in]{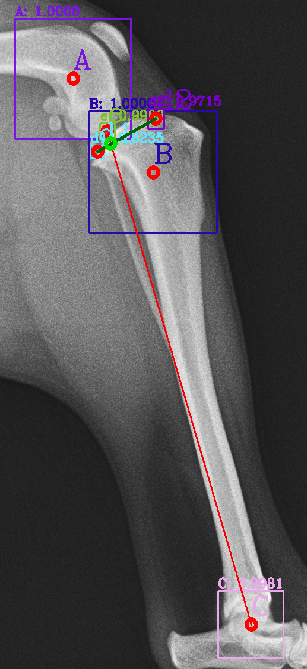}\hfill
    \caption{}
    \end{subfigure}
    \begin{subfigure}[]{0.4\linewidth}
      \centering
      \includegraphics[width=1.2in, height=2in]{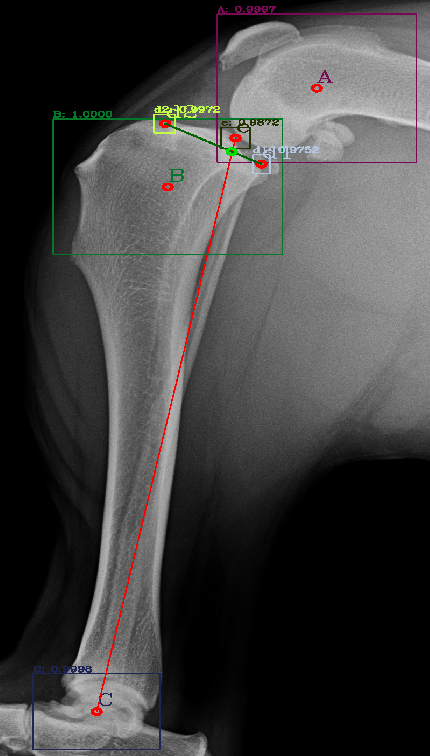}\hfill
    \caption{}
    \end{subfigure}
   \\[\smallskipamount]
   \begin{subfigure}[]{0.4\linewidth}
      \centering
   \includegraphics[width=1.2in, height=2in]{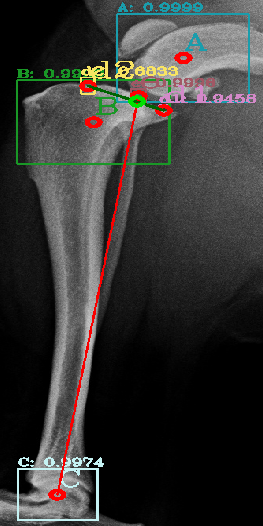}\hfill
   \caption{}
   \end{subfigure}
   \begin{subfigure}[]{0.4\linewidth}
      \centering
      \includegraphics[width=1.2in, height=2in]{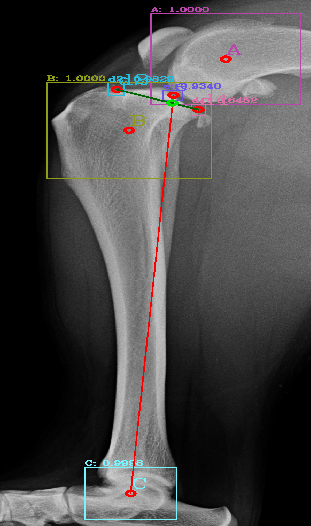}\hfill
   \caption{}
   \end{subfigure}
    \caption{\small{Example of algorithm detecting, highlighting the regions of interests.}}\label{fig:tpapred}
\end{figure}  
\begin{table}[tb]
    \centering
    \caption{\small{TPA shown by trained system with calculations from the detected annotations}}
    \begin{tabular}{|c | c|}\hline
    \textbf{Figure Number} & \textbf{TPA} \\ \hline 
    \ref{fig:tpapred}(a) & 20.537\degree\\ \hline
    \ref{fig:tpapred}(b) & 17.354\degree\\ \hline
    \ref{fig:tpapred}(c) & 19.473\degree\\ \hline
    \ref{fig:tpapred}(d) & 23.369\degree\\ \hline
    \ref{fig:tpapred}(e) & 18.435\degree\\ \hline
    \ref{fig:tpapred}(f) & 19.699\degree\\ \hline
    \end{tabular}
    \label{table:tpares}
\end{table}
\subsection{Activation Functions}
\begin{figure}[tb!]
    \centering
    \begin{subfigure}[]{0.4\linewidth}
    \centering
    \includegraphics[trim={0 0.85cm 4cm 5cm},clip,width=1.2in, height=2in]{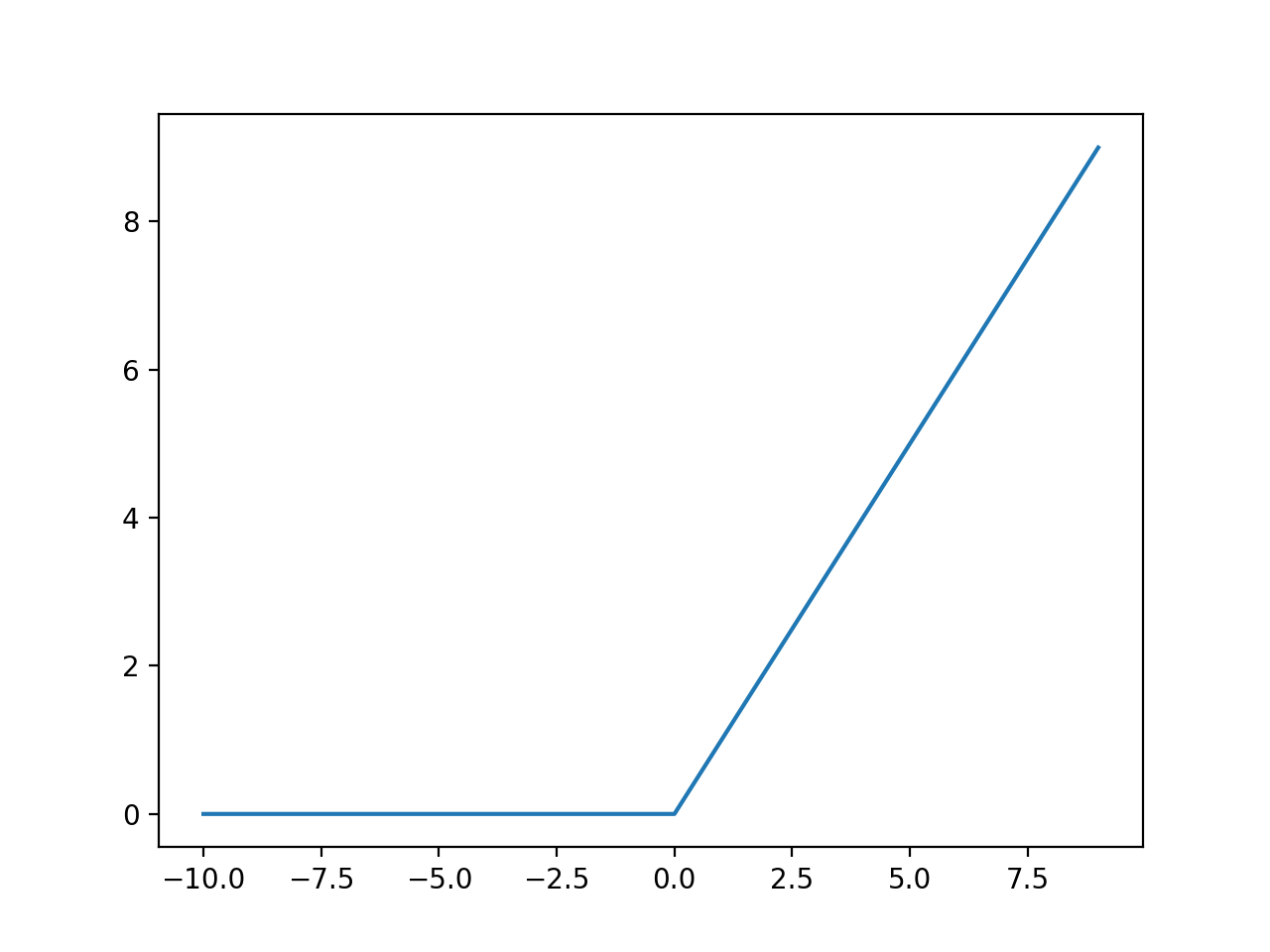}\hfill
    \caption{\small{ReLu}}
    \end{subfigure}
    \begin{subfigure}[]{0.4\linewidth}
    \centering
    \includegraphics[width=1.2in, height=2in]{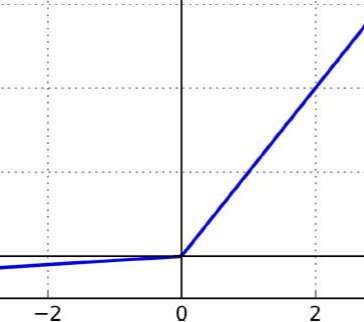}\hfill
    \caption{\small{Leaky Relu}}
    \end{subfigure}\\[\smallskipamount]
    \begin{subfigure}[t]{0.4\linewidth}
    \centering
    \includegraphics[trim={0 0 4cm 5cm},clip,width=1.2in, height=2in]{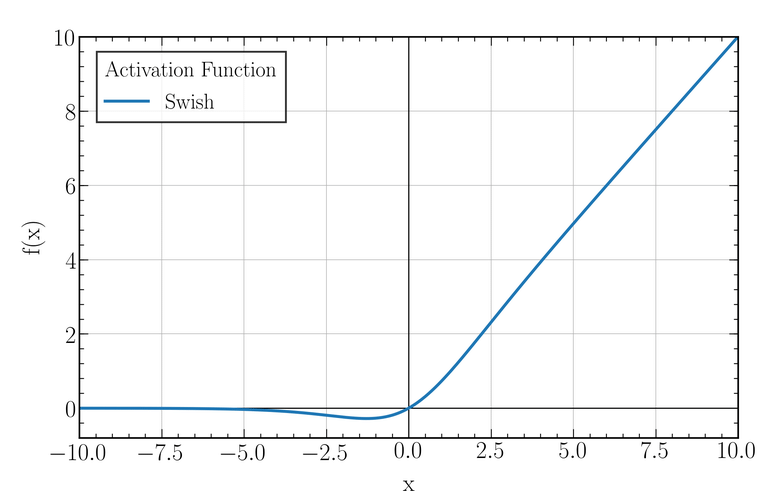}\hfill
    \caption{\small{Swish}}
    \end{subfigure}
    \begin{subfigure}[t]{0.4\linewidth}
    \centering
    \includegraphics[trim={0 0.3cm 4cm 5cm},clip,width=1.2in, height=2in]{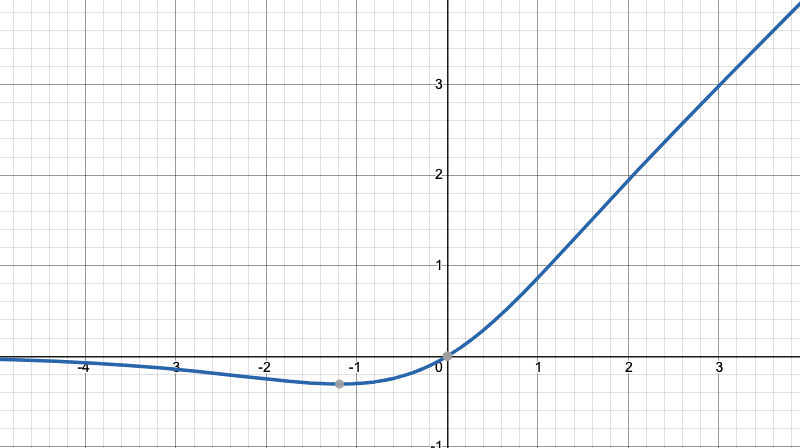}\hfill
    \caption{\small{Mish}}
    \end{subfigure}
    \caption{\small{Graphical form of the activation functions \cite{serengil_2020}.}}
    \label{fig:act}
\end{figure}

\begin{figure}[t]{}
    \centering
    \includegraphics[scale=0.35]{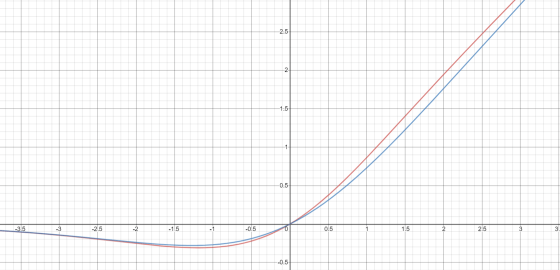}\hfill
    \caption{\small{Mish and Swish comparison for understanding how similar they are. Blue line represent Swish and red represents Mish \cite{serengil_2020}.}}
    \label{fig:actcomp}
    \end{figure}

\noindent As activation functions are explored in this paper for the purpose of performance comparison, a small section has been dedicated for understanding of the roles and types of activation function that have been used. Activation function is simply the pathway that allows feeding of the input and output to and from the current neuron. This can range in variety of forms, from as simple as an on/off switch---i.e, Step function to as complex as the Sigmoid. In this section only the functions used for results obtained are discussed:
\begin{enumerate}
   \item Linear activation function is exactly what the name suggests. It takes an input, multiples it with the learned weight and produces an output that is a function of the input.
   \begin{equation}
    f(x)=mx
    \end{equation}
   \item ReLU is a rectified version of the linear function that does not allow the negative inputs.
   \begin{equation}
    f(x)=max\begin{cases}x & \text {for $x>0$}\\0 & \text {otherwise}
    \end{cases}\end{equation}
    The graphical representation of this is shown in Fig. \ref{fig:act}(a).
    \item Leaky ReLU is a variation of ReLU that has a small slope on the negative area and thus is more complacent toward negative inputs.
    \begin{equation}
    f(x)=max\begin{cases}x & \text {for $x>0$}\\ax & \text { $x<=0$}
    \end{cases}\end{equation}
    The graphical representation of this is shown in Fig. \ref{fig:act}(b).
   \item Swish is a gated sigmoid function that has an interesting mathematical form.
   \begin{equation}
    f(x)=x\times \sigma  (\beta (x))
    \end{equation}
    where the $\beta$ can be a constant or a trainable quantity, depending on which the Swish may act either like a scaled linear (for $\beta=0$) or a ReLU (for $\beta \to \infty$) function. The graphical representation of this is shown in Fig. \ref{fig:act}(c).
    \item Mish is an improvement on the existing swish function; it is a smooth and non-monotonic function that can be defined as:
    \begin{equation}
    f(x)=x\times \tanh (\ln (1+e^x)
    \end{equation}
    The graphical representation of this is shown in Fig. \ref{fig:act}(d). Since Mish and its predecessor are visibly indistinguishable, a comparison of the two is shown in Fig. \ref{fig:actcomp} that displays the negligible difference between the two. 
\end{enumerate}
\section{Testing and Analysis}

\begin{table}[tb]
\centering
\caption{\small{Comparison of results with variation in algorithm versions}}
\begin{tabular}{|c|c|c|c|c|}\hline
     \textbf{Image} & \textbf{YOLOv3} & \textbf{YOLOv4-1} & \textbf{YOLOv4-2} & \textbf{YOLOv4-3} \\ \hline 
    \ref{fig:tpapred}(a) & 20.537\degree & 22.67\degree & 24.8\degree & 22.9\degree\\ \hline
    \ref{fig:tpapred}(b) & 17.354\degree & 17.62\degree & 18.7\degree & 16.22\degree\\ \hline
    \ref{fig:tpapred}(c) & 19.473\degree & 19.53\degree & 20.03\degree & 18.24\degree\\ \hline
    \ref{fig:tpapred}(d) & 23.369\degree & 27.37\degree & 29.03\degree & 25.8\degree\\ \hline
    \ref{fig:tpapred}(e) & 18.435\degree & 15.5\degree & 17.65\degree & 16.8\degree\\ \hline
    \ref{fig:tpapred}(f) & 19.699\degree & 18.24\degree & 20.22\degree & 18.59\degree\\ \hline
\end{tabular}
\label{table:comp1}
\end{table}
\noindent For comparison purposes the radiographs have been tested with YOLOv3, original YOLOv4 \cite{bochkovskiy2020yolov4} and custom modifications of the YOLOv4 by changing the activation functions and the results are shown in Table \ref{table:comp1}. The original version of YOLOv4 (addressed as YOLOv4-1 in Table \ref{table:comp1}) combined the Mish, Linear and Leaky activation functions, the versions YOLOv4-2 and YOLOv4-3 used for the purpose of testing in this paper have combinations of Mish, Linear, Swish and Mish, Linear and Relu activation functions respectively. Similarly, comparison of the results that did not fall under the presumed normal range shown in Fig. \ref{fig:tpapredwrng}, for the algorithm, are shown with variations in activation function in Table \ref{table:comp2}. It can be seen here that these images, are giving TPA predictions similar to the original YOLOv3, i.e. outside of presumed range.
\begin{table}[tb]
\centering
\caption{\small{Comparison of `below range' results with variation in algorithm versions}}\label{table:comp2}
\begin{tabular}{|c|c|c|c|c|}\hline
     \textbf{Image} & \textbf{YOLOv3} & \textbf{YOLOv4-1} & \textbf{YOLOv4-2} & \textbf{YOLOv4-3} \\ \hline 
    \ref{fig:tpapredwrng}(a) & 10.4\degree & 11.22\degree & 10.48\degree & 14.91\degree\\ \hline
    \ref{fig:tpapredwrng}(b) & 10.3\degree & 8.86\degree & 11.3\degree & 11.11\degree\\ \hline
    \ref{fig:tpapredwrng}(c) & 6.53\degree & 8.84\degree & 7.07\degree & 8.18\degree\\ \hline
\end{tabular}
\end{table}    

\section{Conclusion and Future Work}
\begin{figure}[t]
    \centering
    \begin{subfigure}[t]{0.26\linewidth}
    \centering
    \includegraphics[width=0.8in, height=2in]{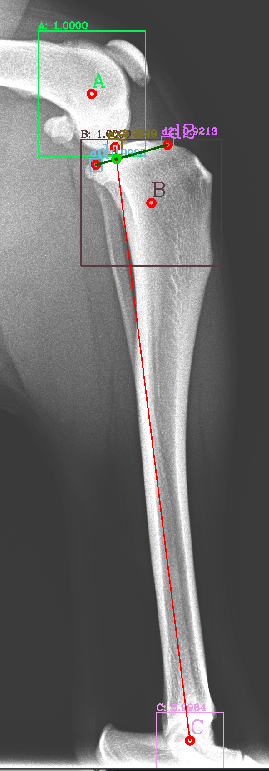}\hfill
    \caption{\small{TPA=10.4\degree}}
    \end{subfigure}
    \begin{subfigure}[t]{0.26\linewidth}
    \centering
    \includegraphics[width=0.8in, height=2in]{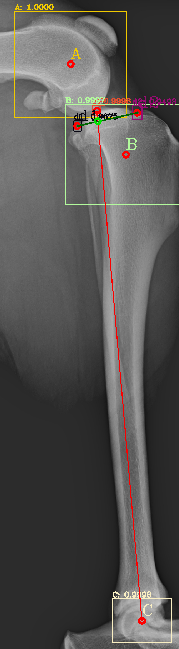}\hfill
    \caption{\small{TPA=10.3\degree}}
    \end{subfigure}
    \begin{subfigure}[t]{0.26\linewidth}
    \centering
    \includegraphics[width=0.8in, height=2in]{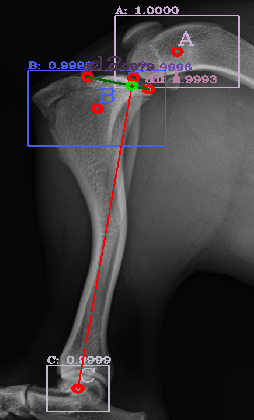}\hfill
    \caption{\small{TPA=6.53\degree}}
    \end{subfigure}
    \caption{\small{Example of algorithm detecting, highlighting the regions of interests that are below known range of value.}}\label{fig:tpapredwrng}
    \end{figure}

\noindent Automated image annotation has become a major requirement in the medical field, since it can be a great tool to drive quick, intelligent and reliable patient care decisions \cite{grady2020system}. Grady and Schaap  \cite{grady2020system}, patented the idea of incorporating user input as part of the learning process, as it is essential that the automated results are validated and appropriately corrected by a user, i.e., a radiographer, when required. 
\indent Results that do not fall under the range mentioned by \cite{seo2020measurement}, are shown in Fig. \ref{fig:tpapredwrng} and Table \ref{table:comp2}. These open a possible scope for improvement and future work as these samples might require human expert-intervention in order to correct the annotations and relearn from the said corrections.
\indent For future work, a user based feedback loop will be added to this system that may be used alongside these predictions, as input to train the system further which will result in more accurate TPA calculation. Another degree to this work could be automating this angle value calculation, i.e., using the TPA values as part of the information fed to the system so that it may be able to draw the lines along the bone joint-axes, and calculate the TPA, all within the same deep neural network; simplifying the user interface to just inserting the X-ray and, potentially resulting in a more accurate TPA calculation.\\
\indent The core finding of this paper is that, even without expert intervention, automation of annotations was successfully performed to a significantly accurate degree, 4 out of every 5 images tested in average. With this in mind, in addition to the previously developed lightweight classifier \cite{Tonima_2021}, it can be concluded that the second step of our development, of the automation in diagnostic tool, is complete. This paper confirms that it is possible to automate the system via annotation which can be improved with the formerly mentioned expert feedback in the future.
\section*{Acknowledgment}
\noindent The dataset used in this paper is provided by Innotech Medical Industries Xray. The research work is financially supported in part by a MITACS Accelerate Project (no. FR56849) under the partner organization Innotech Medical Industries Corp. and the Natural Sciences and Engineering Research Council of Canada (NSERC).

\bibliographystyle{unsrt}
\bibliography{TPA}
\end{document}